# Title: A Personalized Computational Framework for Assessing the Sufficiency of Partially Observed Data in Healthcare AI models.

**Qingchu Jin[1,2], Felistas Mazhude[3], Jamie B. Rabb[3], Robert S. Kramer[3], Douglas B. Sawyer[2,3], Raimond L. Winslow[1,4,5,6,7]**

1. Roux Institute, Northeastern University, ME USA
2. MaineHealth Institute for Research, ME USA
3. MaineHealth, ME USA
4. College of Engineering, Northeastern University, MA USA
5. Khoury College of Computer Sciences, Northeastern University, MA USA
6. Bouvé College of Health Sciences, Northeastern University, MA USA

**Contact information of corresponding author:**

Raimond L. Winslow, PhD

**Email**: r.winslow@northeastern.edu

## Abstract

Achieving early and timely diagnosis and treatment for disease is a major challenge in medicine. Recent applications of machine learning (ML) algorithms trained on patient data have shown promise in many different settings for predicting the time-evolution of patient health state, thereby enabling more reliable patient risk stratification. A challenge often faced when applying these ML algorithms is that at any given time, not all clinical variables (features) needed as input to perform prediction tasks are available. We define the concept of full-feature-capacity (FFC) to refer to prediction performance when such algorithms make use of all features on which they were trained. We then introduce Feature Sufficiency Analysis (FSA) - a computational approach for determining whether a subset of all clinical features needed by an AI model is sufficient to achieve FFC. FSA applies Monte Carlo methods to estimate the underlying distributions of missing variables conditioned on features that are available. FSA provides a patient-specific assessment of whether the existing set of measured features achieves FFC. If yes, then there is no need to acquire further inputs and a ML-based prediction or risk stratification may be generated immediately. We provide two case studies: prediction of need for postoperative prolonged ventilation in patients recovering from heart surgery; 10-year mortality prediction in an outpatient cohort. We show that 86% of patients in the heart surgery cohort and 91% of subjects in the outpatient cohort achieved FFC when using fewer than half of the total number of features on which the learners were trained. We also demonstrate that FSA also provides a clinically interpretable feature-ranking methodology based on prediction sufficiency, identifies intrinsically hard-to-predict patient populations, and has the potential to perform cost-aware optimization for clinical data acquisition. FSA provides a generic computational approach for determining whether incomplete clinical information is sufficient to support trustworthy AI-assisted clinical decision-making, thereby facilitating the prospective deployment of healthcare AI systems across diverse clinical settings.

## Introduction

Early and timely diagnosis and treatment for diseases is one of the major challenges in medicine. From children acute appendicitis[1] to sepsis[2], from acute trauma[3] to lung cancer[4], studies have shown that delayed diagnosis leads to an increase of complications and/or mortality.

There is an emerging body of work on the use of machine-learning (ML) and other methods to learn models from patient data that make early prediction of disease onset or disease progression, improve accuracy of disease diagnosis, and better inform timely choice of therapy[5–8]. Many studies show the impressive extent to which models can help improve disease diagnosis, prediction and treatment[5,6,8–22], including using images to classify cancer[20], using longitudinal ICU data to predict septic shock[15], using waveform data to predict neurological outcome[10], using clinical codes to predict the 3-month risk of pancreatic cancer occurrence, using lab result data to diagnose ovarian cancer[23] and use of multimodal data to predict severity of COVID-19 patients[24]. However, very limited predictive AI tools have been deployed into the real-world settings.

A major challenge in deploying machine learning models in real-world clinical settings is that, at the time a prediction is needed, the complete set of features used for prediction is often unavailable. Missing inputs may arise for many reasons. For example, laboratory tests or imaging results may still be pending. The turnaround time (TAT) for the complete blood count (CBC) test is ~30 mins with the highest urgency[25]. TAT for brain MRI ranges from 40 – 224 mins[26]. Healthcare predictive models require a complete set of input features to generate predictions. When some features are unavailable for a patient in the real-world setting, the standard strategy is to impute the missing values and proceed with prediction. However, imputation alone does not eliminate the uncertainty introduced by the fact that data are missing; it merely estimates what the missing values might be. Thus, the prediction for that patient made with incomplete information may not achieve the reported performance for predictions made with the full feature set. The extent of performance degradation depends on the number of missing features, the importance of those missing features to the model, and the patient-specific context. Currently, there is no widely adopted framework that quantitatively assesses how missing features affect prediction reliability on a patient-specific basis. This presents a significant risk in deploying AI models in the clinic.

This challenge is becoming increasingly important as modern healthcare AI systems incorporate multimodal data sources. While multimodal models improve predictive accuracy[27], they also increase the likelihood that some required inputs will be unavailable at the time a prediction is needed because different data modalities have different acquisition times, costs, and availability. Consequently, an important practical question arises: can a model make a reliable prediction using only the information currently available, or must additional data be obtained before a decision can be made?

Assume that we have a validated healthcare AI model that uses K clinical variables (features) as inputs to generate predictions. We hypothesize that not every patient requires knowledge of all K features for a predictor to make confident decisions. Define prediction performance achieved by an ML model when using all K features on which it was trained to be the Full Feature-Capacity (FFC). We say that an ML decision based on < K features is "confident" for a specific patient if, for that patient, knowledge of the un-measured features does not change the ML model decision.

We present a computational framework called Feature Sufficiency Analysis (FSA), designed to ascertain whether a subset of features is sufficient for an ML model to deliver a prediction with full feature capacity. This framework is developed based on a Bayesian approach for estimating the posterior distribution of missing features with uncertainty analysis to determine the impact of missing features on the model's predictive accuracy. We show with two case studies that this framework has the potential to reduce the time and monetary cost for feature acquisition. Rather than following the imputation-prediction ML paradigm, this framework performs Bayesian-based multiple imputation and then evaluates the necessity to obtain unavailable features for each patient. This tool is model-agnostic, ensuring compatibility across various AI models, and it generates inferences tailored to each patient. Following the concept of FFC prediction, the FSA system provides intuitive tools to perform two fundamental tasks in the ML field. The first is feature importance ranking. Here, we validate FSA-based feature ranking by comparison with widely-used feature ranking methods on our case studies. The second is patient grouping. FSA can identify hard-to-predict patient groups, where prediction performance dramatically drops in this group. In summary, we believe FSA is a principled, easy-to-use and important tool to make healthcare risk models cost-effective while preserving the model performance.

## Results.

### Prediction tasks and baseline AI model.

We consider two clinical prediction tasks to illustrate use of FSA. Task 1 is to predict need for prolonged ventilation in patients recovering from heart surgery[28]. Prolonged ventilation is defined as duration of ventilation greater than 24 hrs. The STS Adult Cardiac Surgery Database (ACSD)[29] was queried to develop a dataset including cardiac surgery cases from a single center over a 10-year period from January 1, 2012 to December 31, 2021 that included 9,238 patients. 12 features (Table. S1) were identified as input for the baseline model. Data were separated into training (70%), validation (10%) and test (20%) sets. A Random Forest (RF) learning algorithm was used to generate a baseline predictive model risk score. If the predicted risk score is greater than threshold (0.285) the model makes a positive prediction. The FFC of this model was 0.88 AUC, 0.67 sensitivity, 0.94 specificity, and 0.94 positive predictive value (PPV).

Task 2 is predicting 10-year mortality in an outpatient cohort from the National Health and Nutrition Examination Survey (NHANES). The cohort included 13,442 outpatients with 35 features available per patient for patients across the United States[30]. The data is publicly available and organized from Erion et al.[31], summarized in Table S2. A RF algorithm was used to develop a baseline predictive model. The FFC of this model was 0.86 AUC, 0.79 sensitivity, 0.79 specificity, 0.92 PPV. Detailed description of the model and features are provided in Methods.

These two tasks were chosen to represent two distinct types of clinical prediction tasks. Task 1 involves acute event prediction in highly monitored ICU units. Task 2 involves long-term (10-year) mortality prediction. These two tasks also differ with respect to prevalence of the positive outcomes (~10% for prolonged ventilation) and～75% for mortality prediction).

**Feature Sufficiency Analysis (FSA):**

We introduce Feature Sufficiency Analysis (FSA). FSA is designed to evaluate whether a FFC prediction is feasible using a subset of patient features. The approach is general and can be applied to any predictive task and ML algorithm. The general approach is illustrated in Fig 1. Assume a baseline ML model that uses $K$ features in $N$ patients to predict a binary outcome. FSA answers the following question - for a new patient with $k'<K$ features available, can this model make an FFC prediction despite the inherent uncertainty introduced by missing features? FSA quantifies the uncertainty of unavailable features by applying MICE: Multiple Imputation by Chained

Equations[32]. MICE is a widely used Bayesian-based multiple imputation method. MICE estimates the posterior distribution of unavailable features using a Monte Carlo method based on the historical data and available features for a given patient. For a new patient, FSA uses the k' available features and the estimated posterior distributions for the K-k' unavailable features. The estimated posterior distributions of the missing features are propagated through the baseline ML model to yield the distribution of the risk score produced by uncertainty in these missing features. The risk score distribution is assessed against the predetermined model threshold on the risk score learned during training. If all samples of the estimated risk score distribution surpassed or are below threshold, then the current set of k' features suffice for the ML model to achieve FFC prediction. In this study, we generated 100 realizations of the risk score using Monte Carlo methods to generate the empirical distribution of the risk score given the uncertainty from missing values. If all 100 realizations are greater or less than the model threshold learned in training, then FFC prediction is defined achieved. In contrast, if all 100 samples from the risk score distribution intersect with the threshold, then additional features are needed to obtain the accurate prediction. A detailed description can be found in Methods.

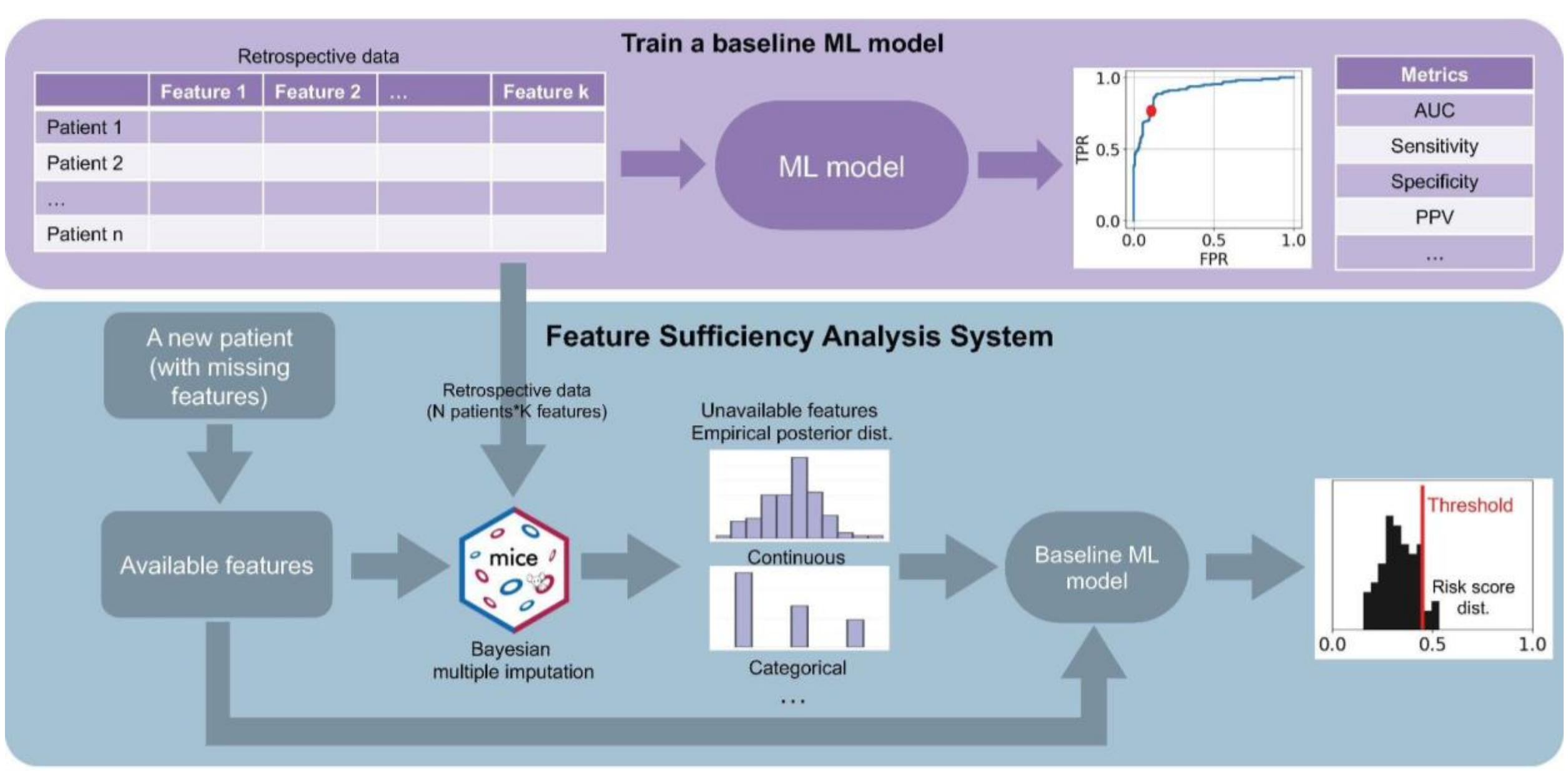

**Figure 1. The diagram of the FSA system.**

**Most patients require <= 50% of all features to achieve FFC prediction.**

One application of FSA is feature importance ranking. FSA evaluates how important a feature is for enabling patients to achieve Full Feature-Capacity (FFC) prediction. To do this, we performed an ablation analysis[33] in which one feature was removed at a time. FSA was then used to estimate the uncertainty introduced by the missing feature and its effect on model predictions for all patients in the hold-out test set ($N$ =1,848 for STS prolonged ventilation, and $N$ = 2,689 for NHANES outpatient mortality). Feature importance was quantified as the percentage of patients (or number of patients) whose risk score distributions intersects the classification threshold after removal of the feature, indicating loss of FFC prediction (Figs. 2a and b). We compared the resulting FSA-based ranking with established methods, including mean decrease in impurity (MDI)[34], permutation importance[34], SHAP values[35], and logistic regression coefficients. The rankings showed broad agreement across methods (Fig. S1 and Table S3). However, unlike conventional metrics, the FSA metric has a direct clinical interpretation: it quantifies the proportion of patients who require a given feature to achieve FFC prediction.

The analyses described above generate a rank-ordered list of feature importance. We then simulate a scenario in which information becomes available incrementally. Starting with only the most important feature, FSA is used to infer the missing features and estimate the corresponding posterior distribution of the model risk score. Next, the two most important features are made available and the analysis is repeated, followed by the top three features, and so on until all features are observed. At each step, the risk score distribution is evaluated to determine whether it supports the same classification as the full-feature model (FFC prediction). The minimum number of observed features required to achieve FFC prediction, denoted k*, provides a patient-specific measure of feature sufficiency. We term this procedure cumulative feature sufficiency analysis (CFSA).

Figs 2c and 2d show the distribution of the minimum number of necessary features k*. Fig. S2 shows the percentage of FFC-predictable patients along with the top *n* important features from FSA-ranking. Notably, ~86% patients for prolonged ventilation task and ~91% patients for 10-year mortality prediction require less than or equal to half of features to achieve FFC prediction.

Figs 2e–j illustrate three representative patient types identified by CFSA. In the first type (Fig. 2e–f), a small number of highly informative features is sufficient to establish that the patient remains below the classification threshold, resulting in rapid achievement of the FFC prediction. In the second type (Fig. 2g–h), a small subset of features is similarly sufficient to establish that the patient remains above the classification threshold. In both types, the risk score distribution remains well separated from the decision boundary despite substantial feature uncertainty. In contrast, the third type (Fig. 2i–j) represents borderline patients whose risk scores lie near the classification threshold. For these patients, uncertainty in the unobserved features substantially affects the model decision, and a much larger number of observed features is required before the FFC prediction can be achieved. These examples demonstrate that prediction complexity is patient-specific: many patients can be confidently classified with only a small subset of features, whereas borderline patients require substantially more information.

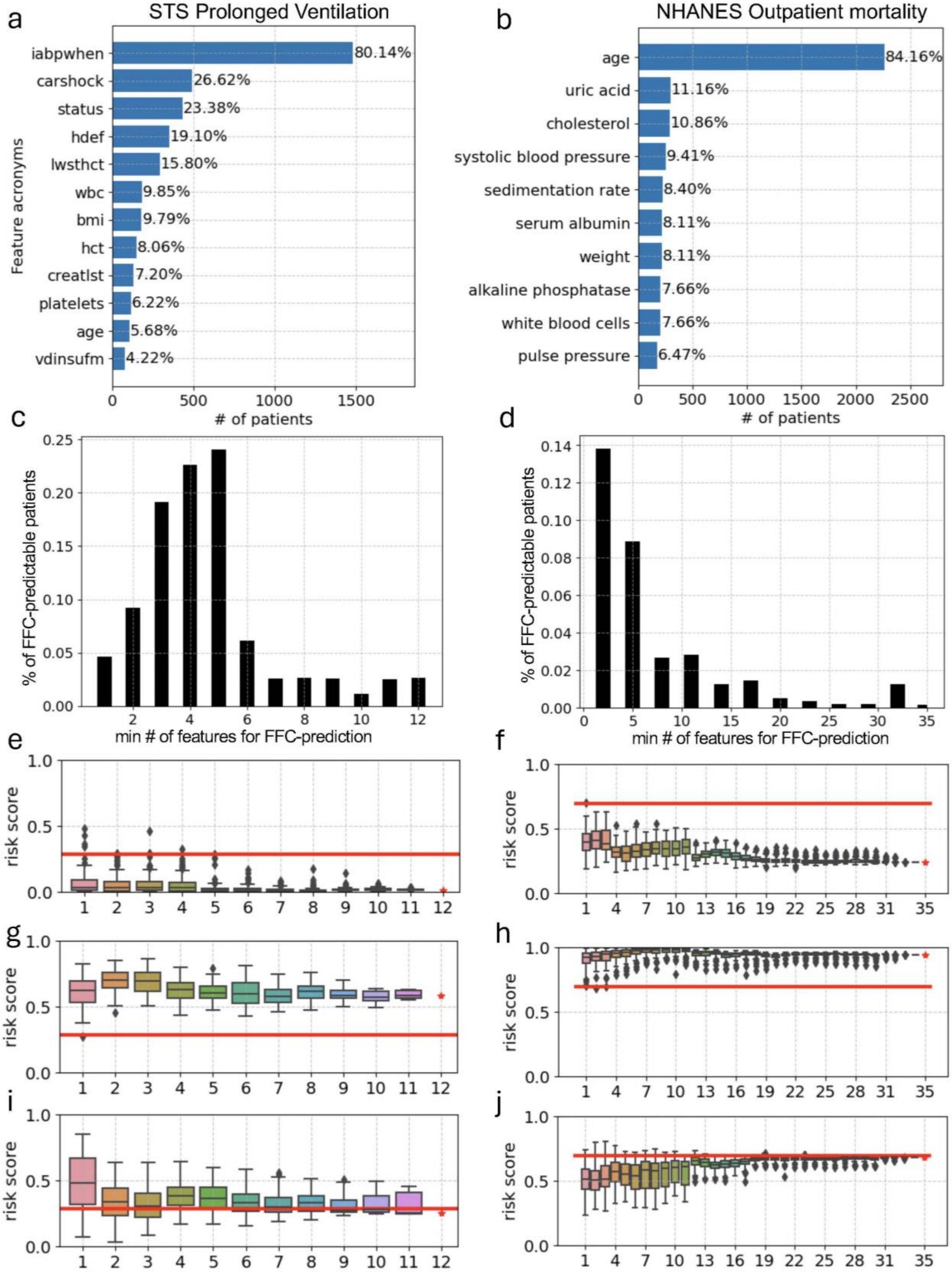

a
STS Prolonged Ventilation
Feature acronyms
iabpwhen
carshock
status
hdef
lwsthct
wbc
bmi
hct
creatlst
platelets
age
vdinsufm
80.14%
26.62%
23.38%
19.10%
15.80%
9.85%
9.79%
8.06%
7.20%
6.22%
5.68%
4.22%
# of patients
b
NHANES Outpatient mortality
age
uric acid
cholesterol
systolic blood pressure
sedimentation rate
serum albumin
weight
alkaline phosphatase
white blood cells
pulse pressure
84.16%
11.16%
10.86%
9.41%
8.40%
8.11%
8.11%
7.66%
7.66%
6.47%
# of patients
c
% of FFC-predictable patients
min # of features for FFC-prediction
d
% of FFC-predictable patients
min # of features for FFC-prediction
e
risk score
f
risk score
g
risk score
h
risk score
i
risk score
j
risk score

**Figure 2. Evaluating the necessity of features for FFC prediction.** Left-side figures are for the prolonged ventilation prediction and right-side figures are for the 10-year mortality prediction. Feature importance is shown for features (top 10 features for mortality prediction) by assessing the number of patients (*n*) and percentage (%) for whom FFC prediction cannot be achieved when specific features are removed (a and b). The distribution of the minimum number of features needed to make FFC prediction is shown (c and d). Boxplots (e - j) for 6 patient examples demonstrate the evolution of risk score distribution with an increasing number of features available. X axis represents the number of available features. Horizontal red line shows the threshold of the ML model and the red star shows the FFC-prediction where all features are available. The inclusion order of features for (c - j) follows the feature ranking in (a and b).

**The FSA strategy yields models that outperform the use of "sub-models"**

One strategy for making predictions in the presence of missing features is to develop a collection of sub-models, where each sub-model is trained on a unique subset of features[31]. Predictions for patients having a particular subset of features available are then made using the appropriate sub-model. However, this strategy of building sub-models becomes combinatorially complex as the number of predictive features increases. In contrast, the FSA system uses one ML model trained with all features, estimates the risk score distribution imposed by the missing features, and determines whether or not FFC prediction can be achieved even in the presence of missing features. To evaluate cohort-level predictive performance in the presence of missing features, we compared FSA with a conventional sub-model strategy. Since FSA generates a risk score distribution for each patient, we used the mean of this distribution as the patient's predicted risk score and calculated the resulting AUC. Fig 3 shows the comparison between FSA method (blue) and sub-model method (red) with the increase of number of available features for both prolonged ventilation prediction (Fig 3a) and mortality prediction (Fig 3b). We iteratively add features based on the feature ranking in Fig 2a and 2b starting from most to least important. These data show that in most cases, FSA outperforms the sub-model method. Notably, FSA demonstrated a smooth and consistent improvement in performance as features were added, whereas the sub-model strategy showed marked fluctuations, especially when predictions were made using only a limited number of features.

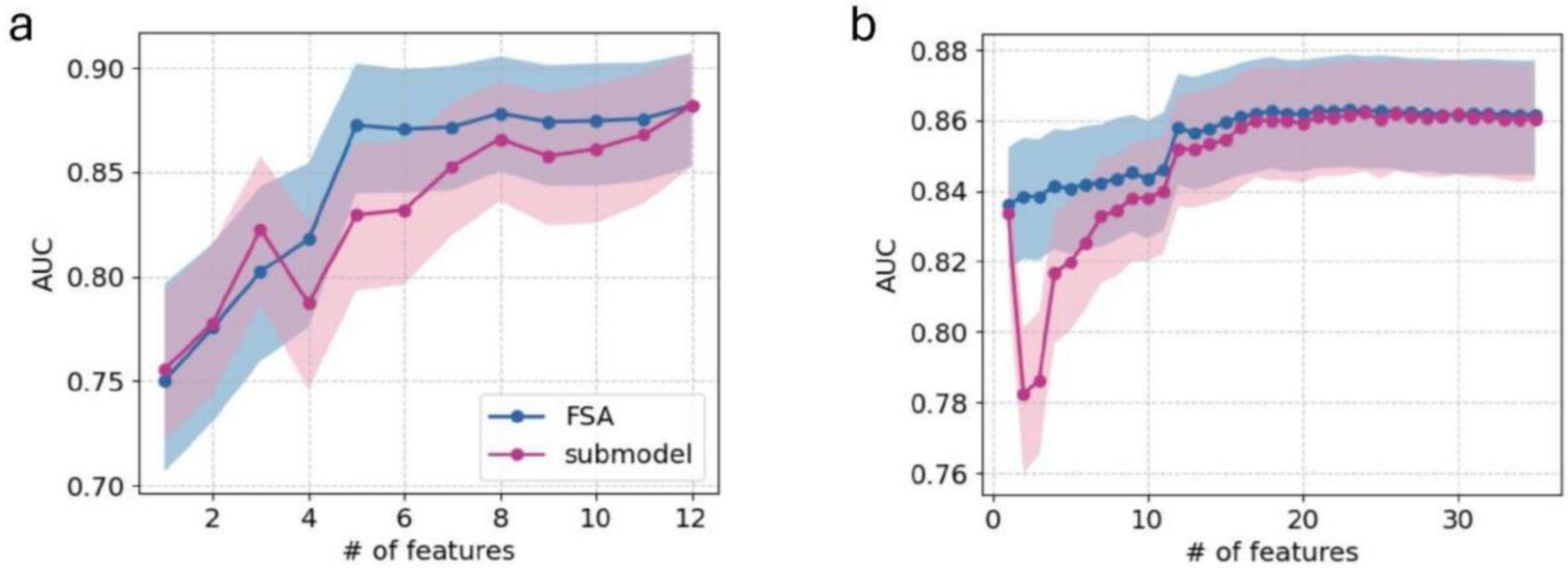

**Figure 3. ML model performance for subsets of features for prolonged ventilation (a) and mortality prediction (b).** AUC (y axis with 95% confidence interval) representing the ML performance changes with the increase of the number of features. FSA method (blue curve) uses the average of risk score distribution to represent the prediction with the performance. Alternatively, sub-models (red curve) are trained for each subset of features and to evaluate the prediction performance.

**Identifying the "hard-to-predict" patient group**

FSA provides a patient-specific measure of the minimum number of features required to achieve Full Feature-Capacity (FFC) prediction (Figs. 2c and 2d). We hypothesized that this quantity could be used to identify patients for whom the model is intrinsically easier or more difficult to predict. We therefore defined easy-to-predict patients as those requiring no more than half of the total feature set to achieve FFC prediction, and hard-to-predict patients as those requiring more than half of the available features.

Applying this definition, the majority of patients belonged to the easy-to-predict group (86% in the STS prolonged ventilation cohort and 91% in the NHANES mortality cohort). As shown in Figs. 4a and 4b, easy-to-predict patients generally had risk scores located farther from the classification threshold, whereas hard-to-predict patients tended to cluster near the threshold, making their predictions more sensitive to uncertainty in missing features.

The distinction between the two groups was associated with substantial differences in model performance. The easy-to-predict group achieved significantly higher AUC than the hard-to-predict group in both prediction tasks (Figs. 4c and 4d). In the NHANES cohort, the hard-to-predict group achieved an AUC of only 0.46 (95% CI: 0.38–0.54), indicating that the classifier performed no better than random guessing in this subgroup. Similar trends were observed for sensitivity, specificity, positive predictive value, negative predictive value, and accuracy (Table 1).

These findings shows that FSA provides a clinically interpretable measure, minimum number of features required for FFC prediction, of prediction difficulty. Furthermore, identification of hard-to-predict patients may highlight populations for whom the current feature set is insufficient and for whom additional clinical variables may be required to improve predictive performance.

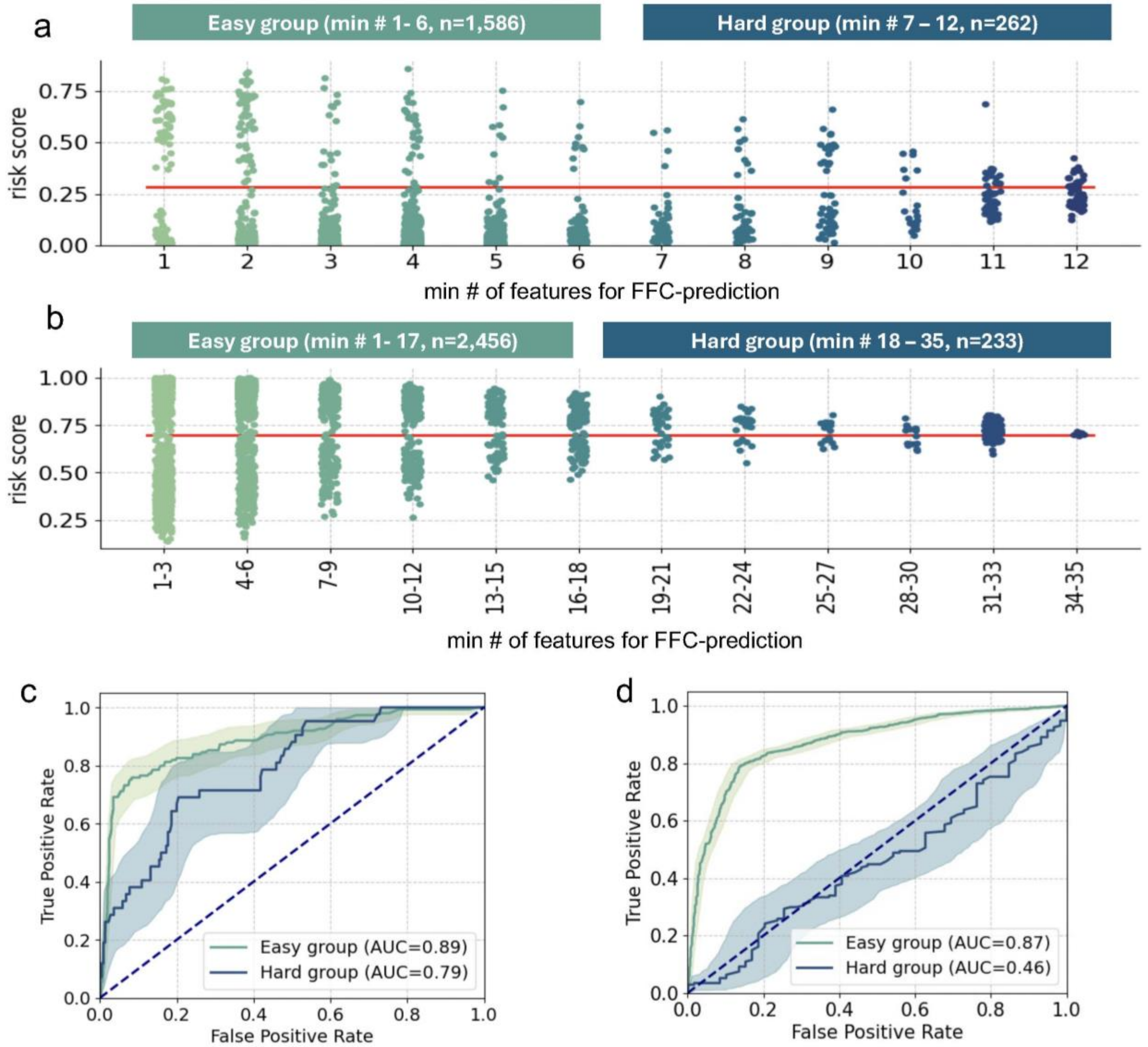

**Figure 4. Grouping patients into easy and hard groups.** For each patient, we identified the minimum number (#) of necessary features for FFC prediction. The dot plot of the risk score categorized by the minimum number of necessary features for FFC prediction is performed for prolonged ventilation prediction on STS heart surgery cohort (a) and 10-year mortality prediction on NHANES cohort (b). We grouped the patients into easy group, defined as # of necessary features < half of total # of features and hard group, defined as # of necessary features > half of total # of features. The horizontal red line is the threshold for binary prediction. The receiver

operating curve with 95% CI is shown in two groups, for prolonged ventilation prediction (c) and mortality prediction (d).

**Table 1. Performance metrics for easy group, hard group and all patients for prolonged ventilation prediction on STS heart surgery cohort and 10-year mortality prediction on NHANES cohort.** Values are presented as mean (95% CI). PPV: positive predictive value, NPV: negative predictive value.

| | **Prolonged ventilation prediction on STS heart surgery cohort** | | |
|---|---|---|---|
| | Easy-to-predict<br>(N = 1,586) | Hard-to-predict<br>(N = 262) | All<br>(N = 1,848) |
| AUC | 0.89 (0.85, 0.92) | 0.79 (0.71, 0.86) | 0.88 (0.85, 0.90) |
| Sensitivity | 0.69 (0.61, 0.77) | 0.60 (0.44, 0.74) | 0.67 (0.60, 0.73) |
| Specificity | 0.96 (0.95, 0.97) | 0.81 (0.76, 0.86) | 0.94 (0.93, 0.95) |
| PPV | 0.65 (0.58, 0.72) | 0.38 (0.26, 0.50) | 0.57 (0.50,0.64) |
| NPV | 0.97 (0.96, 0.98) | 0.91 (0.87, 0.95) | 0.96 (0.95, 0.97) |
| Accuracy | 0.94 (0.92, 0.95) | 0.78 (0.73, 0.82) | 0.91 (0.90, 0.93) |
| | | | |
| | **10-year mortality prediction on NHANES cohort** | | |
| | Easy-to-predict<br>(N = 2,456) | Hard-to-predict<br>(N = 233) | All<br>(N = 2,689) |
| AUC | 0.87 (0.86, 0.89) | 0.46 (0.38, 0.54) | 0.86 (0.84, 0.98) |
| Sensitivity | 0.81 (0.79, 0.82) | 0.62 (0.55, 0.69) | 0.79 (0.77, 0.81) |
| Specificity | 0.84 (0.81, 0.86) | 0.29 (0.17, 0.41) | 0.79 (0.76, 0.82) |
| PPV | 0.94 (0.92, 0.95) | 0.72 (0.65, 0.79) | 0.92 (0.90,0.93) |
| NPV | 0.59 (0.56, 0.63) | 0.20 (0.12, 0.30) | 0.56 (0.53, 0.59) |
| Accuracy | 0.81 (0.80, 0.83) | 0.54 (0.48, 0.60) | 0.79 (0.77, 0.80) |

## Discussion

Early and timely diagnosis and treatment remain one of the major challenges in healthcare delivery. Recently, AI models have shown promise in supporting early diagnosis, detection and treatment in various healthcare applications. However, these AI models often require significant numbers of features as inputs. This in turn can lead to a delay in decision making due to the need to acquire missing data. We hypothesized that not every patient requires all features to make confident prediction. In this study, we developed a Bayesian-based computational framework to quantitatively identify the necessity of features for making confident prediction in a personalized manner.

Our system, Feature Sufficiency Analysis (FSA), is a simple, model-agnostic, Bayesian-based, and personalized computational framework designed to estimate the feature sufficiency for confident prediction. To define the confident prediction, we introduce the concept of full-feature-capacity (FFC) prediction, which refers to ML prediction performance when using all features. A confident prediction for a patient only a subset of feature available means that, regardless of the values of missing features, the ML prediction will remain the same. We refer to this patient as an FFC-predictable patient. For prolonged ventilation prediction, 86% patients only need half of the full complement of model features to achieve FFC prediction. 75% patients reach FFC prediction in less than half of the full data acquisition time and monetary cost. For 10-year mortality prediction, 91% patients reach FFC prediction with half of the full complement of features. Additionally, leveraging the FSA system and the concept of reaching FFC prediction, we developed a FSA-based feature ranking method and a patient grouping method to identify hard-to-predict patients. Both methods are developed based on the minimum number of necessary features for FFC prediction.

One practical application of FSA is to support cost-aware deployment of healthcare AI models. FSA can help to reduce the time and monetary cost, while maintaining FFC prediction when using healthcare ML model To illustrate this application, we performed a proof-of-concept cost analysis using both the STS cohort and the NHANES outpatient cohort. The time and monetary cost of features have been annotated by 2 clinical practitioners for STS prolonged ventilation prediction (Table S4). Features are grouped by the clinical test/actions. We performed a cost analysis, similar to cumulative feature sufficiency analysis, except that we iteratively include clinical tests/actions that may provide data needed to specify multiple features (e.g., results of an echo ultrasound can be used to determine both hdef (ejection fraction) and vdinsufm (Mitral valve insufficiency/regurgitation). Rather than showing the first *n* features, cumulative time/monetary cost of available clinical test/action are shown in the plot. A detailed description of cost analysis can be found in Methods. Cost analysis is based on the cost order of clinical test/action from least to the most. Fig. 5a shows the percentage of FFC-predictable patients as a function of time to perform the clinical test/action. Fig. 5b shows the same as a function of monetary cost. Fig. 5 shows that ~75% patients achieve FFC prediction without the need for an echo ultrasound exam, a relatively costly procedure in terms of time and money to measure features that aren't needed for reliable prediction. A similar conclusion was observed in the monetary cost analysis for 10-year mortality prediction in the NHANES cohort, where 89% of patients achieved FFC prediction using only half of the total monetary cost (Fig. S3). These results indicate that significant time and monetary cost for acquiring clinical features can be potentially saved by our FSA system.

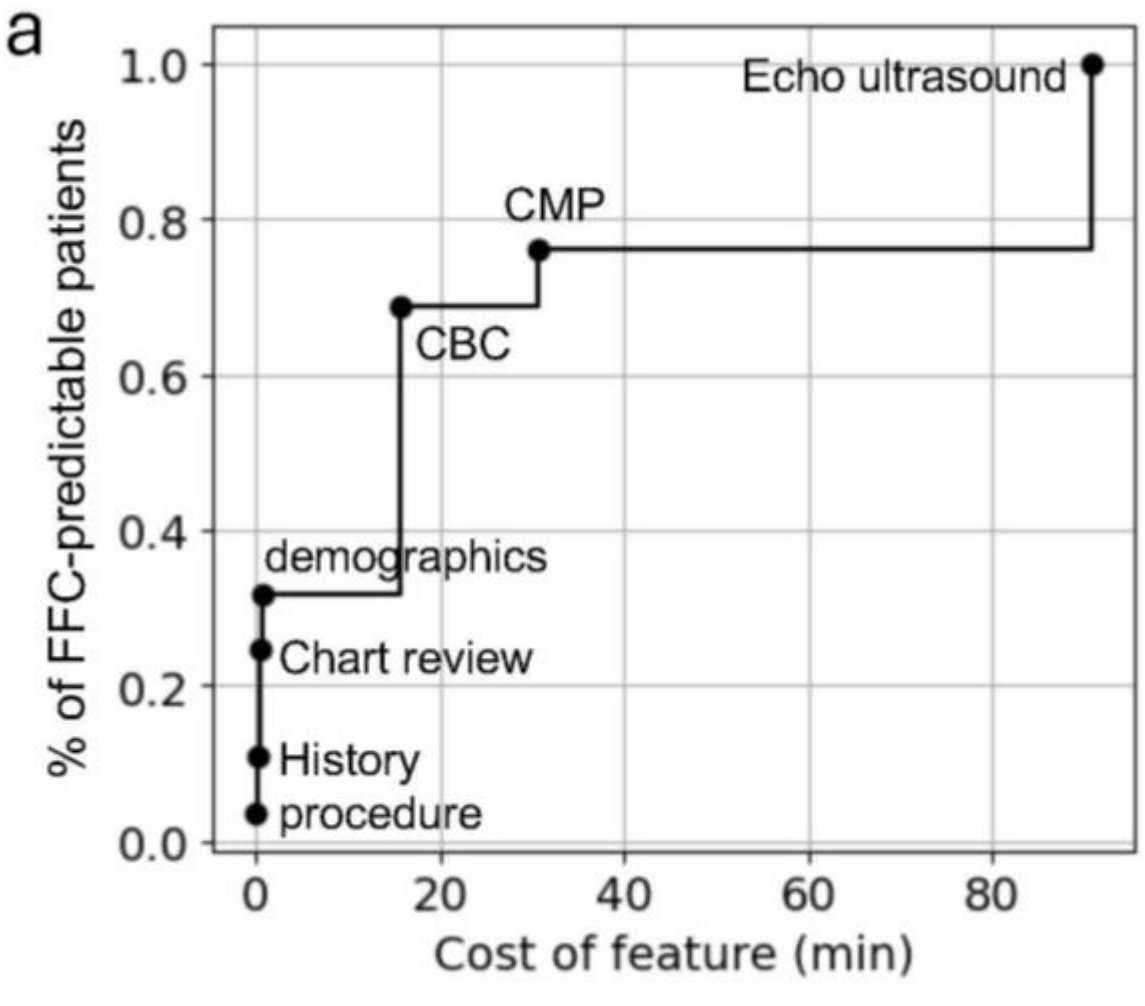

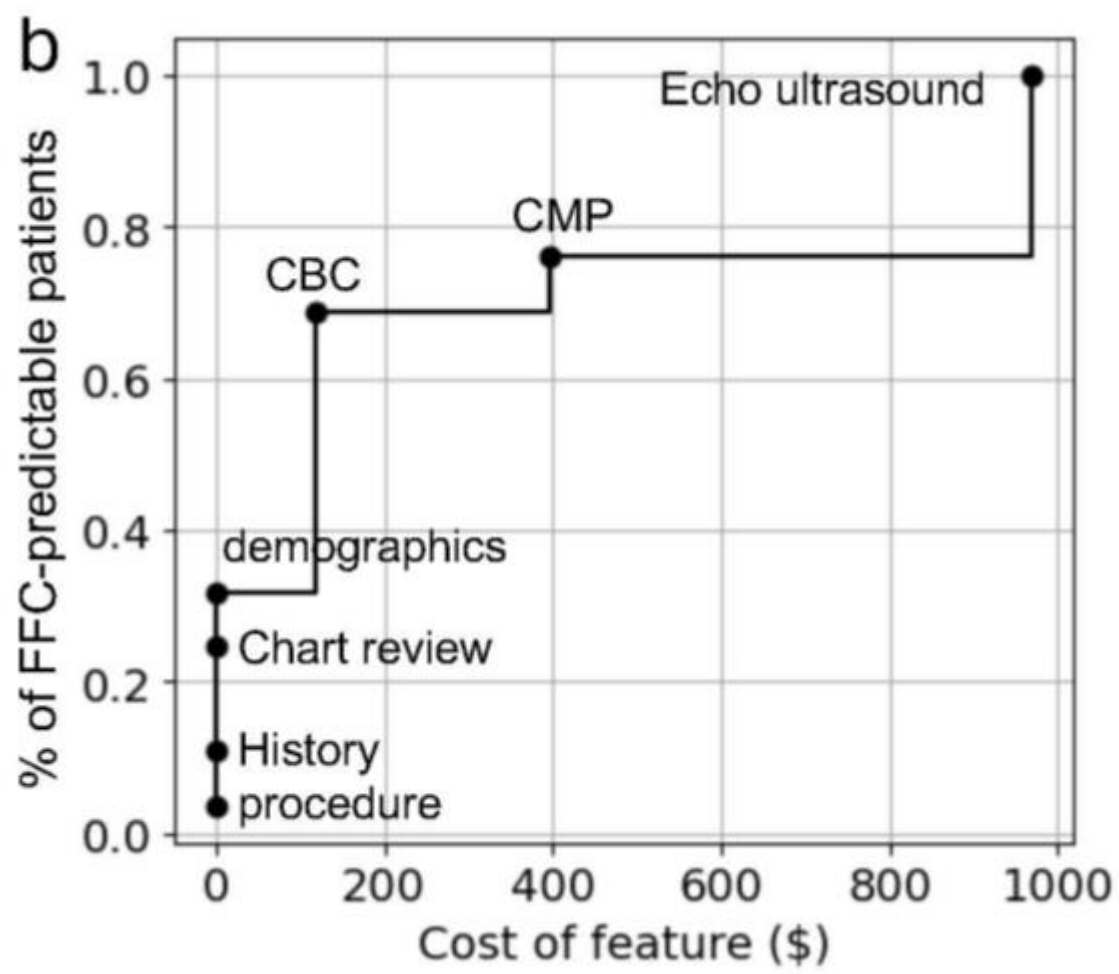

**Figure 5. Time (a) and monetary (b) cost analysis on the prolonged ventilation prediction.** By iteratively adding clinical test/actions to the model, the x axis shows the cumulative time (a) and monetary (b) cost, y axis shows the percentage of patients that can reach FFC prediction given the cost. The clinical test/actions are iteratively added based on the cost order from the least to the most. Each clinical test/action is composed of one or more features.

Our system provides a quantitative recommendation for one of the most common daily jobs of clinicians: whether or not it is necessary to perform an extra test or procedure on their patients before a reliable prediction can be made regarding patient health state. The system is principled, highly intuitive and explainable such that the derived tools including feature ranking and patient grouping can be easily understood. This system is a model-agnostic framework such that all healthcare ML models, even non-ML models, can be applied. Finally, we show that in cases where we must make predictions for patients regardless of whether they are reaching FFC prediction, the FSA-based risk score performs better than does the sub-model method.

One premise underlying the analyses presented in this study is the accurate estimation of posterior distributions for missing clinical variables. To evaluate this assumption, we performed a validation study on the STS heart surgery cohort using the 12 clinical features included in the baseline model. For each feature, values were masked in the test set and imputed using MICE 100 times to generate patient-specific posterior distributions. For continuous variables, we constructed posterior credible intervals of varying sizes (1%–99%) and calculated the proportion of patients whose true observed values fell within their corresponding credible intervals. Accurate posterior estimation would be expected to yield agreement between the credible interval size and the observed proportion of patients contained within that interval, a standard assessment of coverage calibration. For categorical variables, calibration plots were generated by comparing predicted probabilities with observed frequencies for each category. As shown in Fig. S5, the predicted and observed values closely aligned across both continuous and categorical variables, demonstrating accurate estimation of posterior distributions and supporting a key assumption of the Feature Sufficiency Analysis framework.

Identifying hard-to-predict patient groups for whom the model is less predictive is important in the clinical settings. A major effort in ML in healthcare is to quantify prediction uncertainty[37]. Prediction uncertainty (PU) is the distribution associated with the risk score for each patient[38]. We tested the hypothesis that patients with high PU are correlated with our hard-to-predict group. We applied forestCI[39], a widely used PU estimator for random forest learning algorithms, to obtain the PU. For each prediction tasks, we take the top *n* patients with the highest PU. Figs. S4a and S4b show that the hard-to-predict group based on FSA does not overlap strongly with the high PU group. Figs S4c and S4d demonstrate that the high PU groups has an AUC 0.86 (0.80 – 0.90) for prolonged ventilation prediction and 0.75 (0.69 – 0.81) for 10-year mortality prediction. Performance for both groups is better than for the hard-to-predict group defined using FSA, AUC 0.79 (0.71, 0.86) for prolonged ventilation prediction and AUC 0.46 (0.38, 0.54) for 10-year mortality prediction. This suggest that the FS approach identifies a group that is more enriched with hard to predict patients.

We think there are three future research directions of this work. First, for a patient with only a portion of features available and that does not reach FFC prediction, can we identify which the optimal next feature(s) to measure such that the patient has the largest likelihood to reach FFC prediction? This is known as dynamic feature selection. Multiple methods have been developed to do this including using reinforcement or greedy selection[40]. We will explore dynamic feature selection methods using the FSA approach. Second, we will investigate how measurement waiting times impact on FFC prediction. While waiting for additional features to be measured, values of already measured features such as lab results and vital signs may also change, thus creating an additional source of uncertainty in prediction. Additionally, we will explore strategy to identify approaches to improve the model performance in the "hard-to-predict" patient subgroups identified by the FSA methodologies.

We acknowledge some limitations of this study. The cumulative FSA analysis presented in Fig. 2 assumes that features become available according to their predictive importance, with the most informative features added first. Consequently, the finding that most patients achieve FFC prediction with half or fewer of the available features represents an idealized scenario. In real-world clinical settings, where missingness may occur randomly or according to workflow constraints, a larger number of observed features may be required on average to achieve FFC prediction. Nevertheless, these results demonstrate that complete feature availability is not necessary for many patients to achieve the predictive capacity of the full model. Furthermore, FFC prediction should not be interpreted as a guarantee of prediction correctness. Rather, it indicates that the prediction generated from a partially observed feature set is equivalent to using the complete feature set. The accuracy of FFC prediction therefore remains fundamentally constrained by the performance of the underlying full model, which may still produce incorrect predictions.

In summary, FSA has demonstrated a strong capacity to benefit the application of healthcare ML models in practice. It is a principled, easy-to-use system such that can be easily applied to existed clinical score models. FSA can significantly reduce the time and monetary cost for feature collection, while preserving model performance. We believe FSA will be a life-saving tool in the clinical settings.

## Methods

**Datasets.** *STS Heart surgery cohort* One heart surgery patient cohort was obtained by querying the STS Adult Cardiac Surgery Database (ACSD)[29] to develop a dataset including cardiac surgery cases from the MaineHealth (MH) Maine Medical Center over a 10-year period from January 1, 2012 to December 31, 2021. Data were harmonized across the various versions of the STS ACSD. All patient identifiers and private health information (PHI) were removed for patient protection. The project was submitted to the MH Maine Medical Center (MMC) Institutional Review Board (IRB), who determined the project to be "non- research" in a letter dated September 11, 2021.

*NHANES cohort* National Health and Nutrition Examination Survey (NHANES) is a publicly-available US national outpatient cohort of 13,442 patients and 35 features[30] from 1971 to 1974. The 10-year mortality status was followed up in 1992.

**Time and Monetary cost.** The time cost for STS features were assessed by two clinical practitioners at the MH MMC and presented in Table S4. Variables are grouped based on the Test/Action it requires to obtain. For each Test/Action, the time cost is provided in a range. In this study, we used the minimum value of the range to represent the time cost. The monetary cost for STS variables is obtained from Hospital Price Index[40]. Note that one test can generate multiple variables. For example, Echo ultrasound provides both vdinsufm and hdef. Thus, cost analysis on the ML model in Fig. 5 is based on Test/Action rather than variables. Erion et al.[31] provided the monetary cost for the 35 features in NHANES dataset, summarized in Table S5. They assigned costs to features by referencing the Medicare data for lab tests (Clinical Laboratory Fee Schedule Files - Cy 2019 Q3 Release (Centers for Medicare and Medicaid Services,2019);

https://cms.gov/Medicare/Medicare-Fee-for-Service-Payment/ClinicalLabFeeSched/Clinical-Laboratory-Fee-Schedule-Files.html).

**Baseline ML model.** The raw data are imputed by the mean value for the continuous variable and mode value for the categorical variables. The data is stratified randomly sampled into training (70% for STS dataset and 60% for NHANES dataset), validation (10% for STS dataset and 20% for NHANES dataset) and test (20%) set.

The aim of the ML model for the heart surgery data is to predict the occurrence of postoperative prolonged ventilation. To do this, we collected the 72 expert-selected features identified by O'Brien et al [41] to predict prolonged ventilation. We developed a random forest model on 72 features having 0.89 AUC. We identified the 12 most important features by feature ranking using mean decrease of impurity[34]. The 12-feature model yields 0.88 AUC, 0.67 sensitivity, 0.94 specificity, 0.94 positive predictive value (PPV) and the threshold is 0.285. We chose the baseline ML model as the 12-feature random forest model for the prediction of the prolonged ventilation. 12 features are shown in the Table S1.

The task for NHANES dataset is to develop a ML model to predict 10-year mortality with 35 features. We also developed a random forest model on 35 features with 0.86 AUC, 0.79 sensitivity, 0.79 specificity, 0.92 positive predictive value (PPV), 0.56 negative predictive value (NPV). The threshold of the risk score is 0.696.

**Feature Sufficiency Analysis (FSA).** Feature Sufficiency Analysis (FSA) is a Bayesian-based computational framework. The system aims to identify the effect of missing variables on the binary prediction. Using a baseline ML model, we define a prediction made with a complete set of features as the full-feature capacity (FFC) prediction. For a patient with only a subset of features available, if the prediction remains the same regardless of the values of the rest of missing variables, we consider that the subset of features for this patient is sufficient, reaching the full-feature capacity (FFC) prediction. This indicates that the decision made with the available subset of features remains unchanged even when considering all possible values for the missing features. Namely, missing variables do not affect the binary prediction.

FSA analyzes the impact of missing values on prediction through a three-step process.

1. *Posterior distributions for missing values*. For each patient, we perform a Monte Carlo approximation to estimate the posterior distributions for missing values conditional on the available features, represented as *P(missing variables | available variables)*. We applied the Multiple Imputation with chained equations, a widely-used numerical algorithm to estimate the posterior distribution for missing values. This is a distribution-free Monte Carlo algorithm for both numerical and categorical variables. In this study, we generated 100 Monte Carlo realizations to approximate these posterior distributions.

2. *Risk score calculation*. Each of the 100 Monte Carlo realizations, combined with the available feature values was put into the baseline ML model to obtain the corresponding risk scores, thus we obtained 100 risk scores from 100 realizations in step 1.

3. *Risk score distribution analysis.* 100 risk scores form a risk score distribution driven by the posterior distributions of missing values. By examining the relative position between risk score distribution and the threshold of ML model, we could assess the effect of missing values on prediction. If all 100 risk scores exceed the threshold, it indicates that, regardless of the missing values, the ML model will make a positive prediction, and vice versa. We consider this patient with a subset of features available reaching the FCC prediction. If some risk scores are above and some are below the threshold, it indicates that missing variables can affect the prediction. In other words, the risk score distribution intersects with the threshold. We consider that this patient requires missing variables to reach FCC prediction.

**Feature ranking based on FSA.** We performed feature ablation studies to evaluate feature ranking based on FSA. Ablation study assesses feature importance by evaluating how removing/ablating this feature affects model performance. In this study, we remove the feature to be assessed from the test set and set it to be missing. Then, we ranked the feature importance by the number of patients for whom FCC prediction cannot be achieved with this specific feature being removed. Greater number of patients indicates a greater importance for this feature.

**Comparison with other feature ranking methods.** We compared our FSA feature ranking with 4 widely used feature ranking methods in the ML community: mean decrease of impurity[34], permutation[34], logistic regression and SHAP[35]. The Spearman correlation is performed to show the pairwise correlation between different feature ranking methods.

**Validation for the posterior distribution of missing variables.** A faithful estimation of the posterior distribution for missing values is an essential step for the FSA system. We validated the posterior distribution by leveraging the posterior distribution from the ablation study for feature ranking. For continuous variables, we set a credible interval on the posterior distribution for all patients in the test set, centered at 50%. For example, a 90% credible interval ranges from 5% to 95% of the posterior distribution. We then count the number of patients whose actual variable values fall within this credible interval range. The percentage of patients falling within the range should match the credible interval range. To validate the posterior distribution, we vary the credible interval range from 1% to 99% in 1% increments and calculate the corresponding percentages of patients falling within the credible interval. For the categorical variables, we performed the calibration plot on the test set data for each unique value of the variable. The calibration plot shows the predicted probability of a value for the categorical variable from the posterior distribution and the percentage of patients having that specific value.

**Cumulative FSA (CFSA).** With a machine learning predictor with $K$ features, we first define an order of the feature. The order can be based on the feature importance or the time/monetary cost of features. Then, we make first $n$ features available based on the order and let the FSA system identify the risk score distribution for the first $n$ features. We enumerate $n$ from 1 to $K$. For one patient, it enables us to analyze the evolution of the risk score with the increase of $n$ and identify the minimum necessary number of features to reach FFC prediction given the feature order. For a patient cohort, we can identify the percentage of patients who can reach FFC prediction with first $n$ features.

**Cost analysis.** For the STS heart surgery cohort, features are grouped based on the clinical test/actions to share the time and monetary cost. Estimates of time and monetary costs for each clinical test/procedure were generated by a clinical research coordinator at Maine Medical Center based on the experience and the publicly available pricing information from CompareMaine[42], a statewide healthcare cost transparency resource for hospitals. The time and monetary cost have been summarized in Table S4. When a clinical test was associated with a range of time cost, the minimum reported time was used in the analysis. In the NHANES cohort, each feature assigns a monetary cost described in Erion et al[31], summarized in Table S5. Similar to cumulative feature sufficiency analysis, we form an order from low to high cost and identify the percentage of FFC-predictable patients with first *n* features based on the order of cost. For the STS heart surgery cohort, since features are grouped based on clinical test/actions, rather than first *n* features, we use first *n* clinical test/actions to perform the analysis.

**ML performance with missing variables.** FSA system is able to provide a risk score distribution for patients with missing variables. We took the mean of the risk score distribution from FSA and obtained the prediction by comparing the mean risk score with the threshold. For comparison, another strategy for dealing with missing values is to develop a collection of sub-models, where each model is based on a unique subset of features[31]. Thus, prediction with missing variables only needs to identify the right sub-model that is trained with available features.

Based on the feature importance order, we make first n features available and evaluate the ML performance by AUC. We enumerate *n* from 1 to K to analyze how AUC evolves with more features being available for both FSA system method and sub-model method.

**Patient grouping based on the FSA system.** Following the feature importance order, we performed the cumulative feature sufficiency analysis to identify the minimum number of features necessary to reach FFC prediction for every patient. We define patients that require less or equal to half of the total number of features to reach FFC prediction as "easy group", and those who require more than half of total number to reach FFC prediction as "hard group". We compare two groups of patients by evaluating the ML performance on both groups.

## Data Availability

The NHANES outpatient cohort is publicly available from Erion et al.[31] and has been integrated into the code repository submitted with this manuscript for peer review. The STS cardiac surgery cohort is not publicly available due to patient privacy concern.

## Code Availability

The source code implementing Feature Sufficiency Analysis (FSA) is provided as a supplementary software file accompanying this manuscript for peer review. Due to patient privacy and institutional data-sharing restrictions, the code for the STS prolonged ventilation prediction cannot be fully executed without access to the original clinical dataset. The code for the NHANES 10-year mortality prediction is fully reproducible using the publicly available NHANES dataset. Upon acceptance of the manuscript, the FSA source code will be made publicly available through a GitHub repository.

## Acknowledgement

We would like to thank for the funding support of the HEART Impact Engine of the Northeastern University. Drs. Robert Kramer and Douglas Sawyer receive funding from NIH/NIGMS UM1Hl147371, Dr. Douglas Sawyer is supported by NIH/NIGMS P20GM139745.

## Conflict of Interest

Authors have filed one patent application related to this work: application no. US19/288,506 on the system and method for feature-based machine learning model prediction (inventors: R.L.W. and Q.J.)

# Supplementary information for

# A Personalized Computational Framework for Assessing the Sufficiency of Partially Observed Data in Healthcare AI models.

**Qingchu Jin[1,2], Felistas Mazhude[2], Jamie B. Rabb[2], Robert S. Kramer[2], Douglas B. Sawyer[2,3], Raimond L. Winslow[1,4,5,6]**

1. Roux Institute, Northeastern University, ME USA
2. MaineHealth Institute for Research, ME USA
3. Maine Medical Center, ME USA
4. College of Engineering, Northeastern University, MA USA
5. Khoury College of Computer Sciences, Northeastern University, MA USA
6. Bouvé College of Health Sciences, Northeastern University, MA USA

## Contents

### Supplementary Figures

### Supplementary Tables

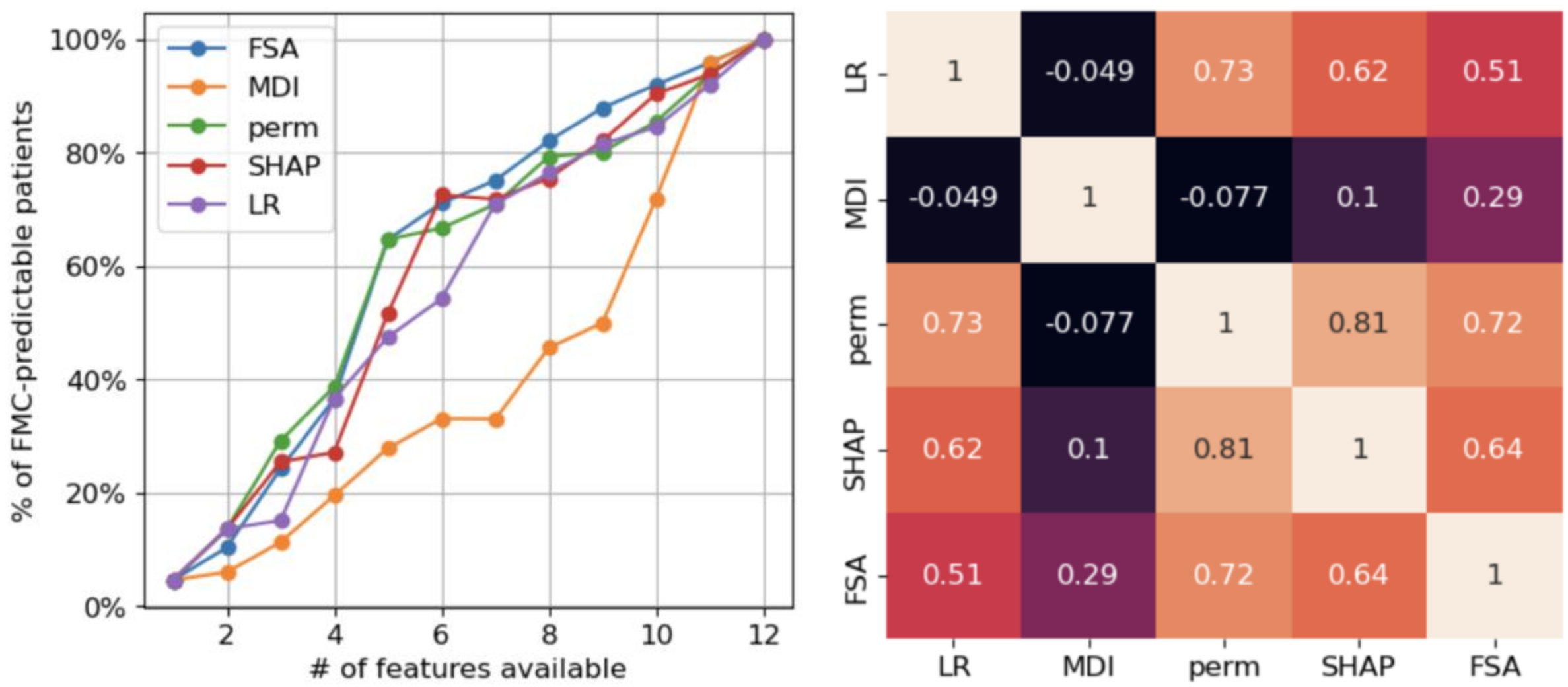

**Figure S1. Comparison between FSA-based feature ranking with 4 widely-used ranking methods.** In left figure, we performed the cumulative feature sufficiency analysis for the STS prolonged ventilation prediction based on 5 different feature order including the order of FSA-based feature ranking and 4 widely used ranking methods: mean decrease of impurity (MDI), permutation ranking method, Shapley value and logistic regression. The feature order for the cumulative feature sufficiency analysis is based on the feature ranking described in Fig 1a. X axis is the first n features available based on the feature ranking order. Y axis is the % of FFC-predictable patients. The right figure is a pairwise spearman correlation between different ranking methods.

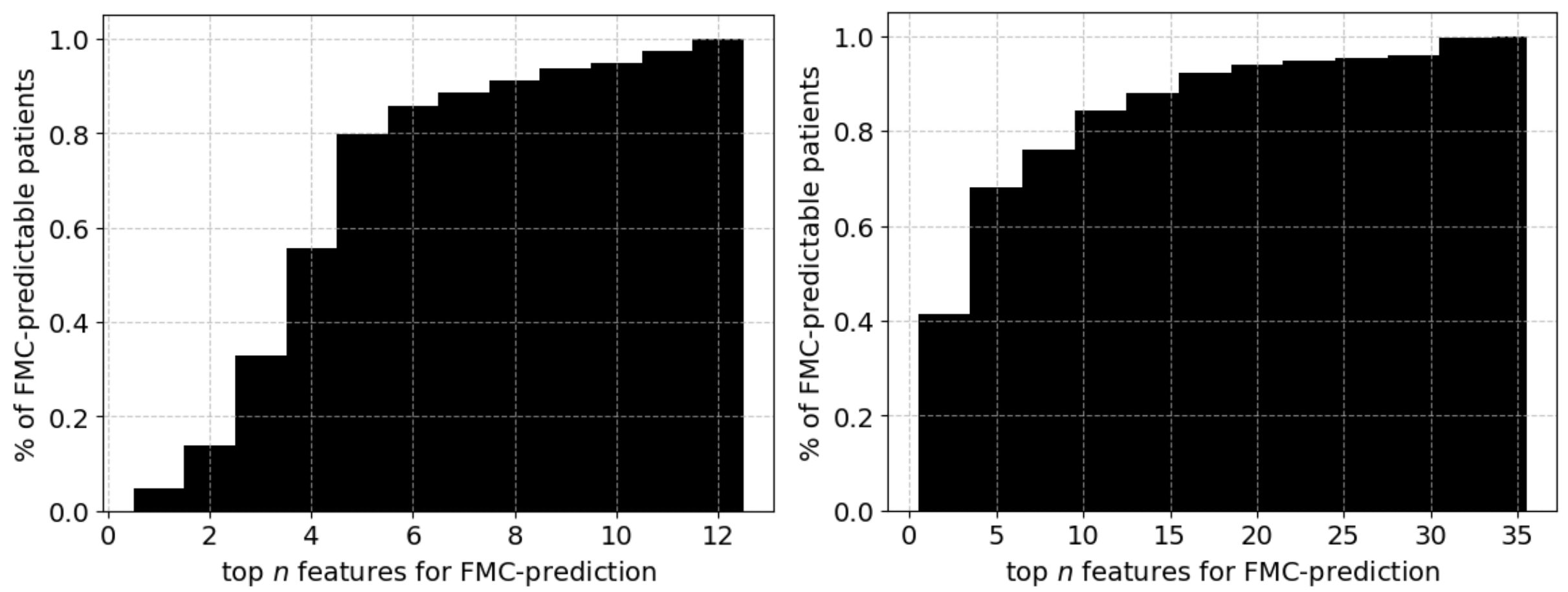

**Figure S2. Cumulative feature sufficiency analysis for prolonged ventilation (left) and mortality prediction (right).** Following the FSA-based feature ranking, we identify the % of FCC-predictable patients with the top *n* features based on the ranking from the most to least important.

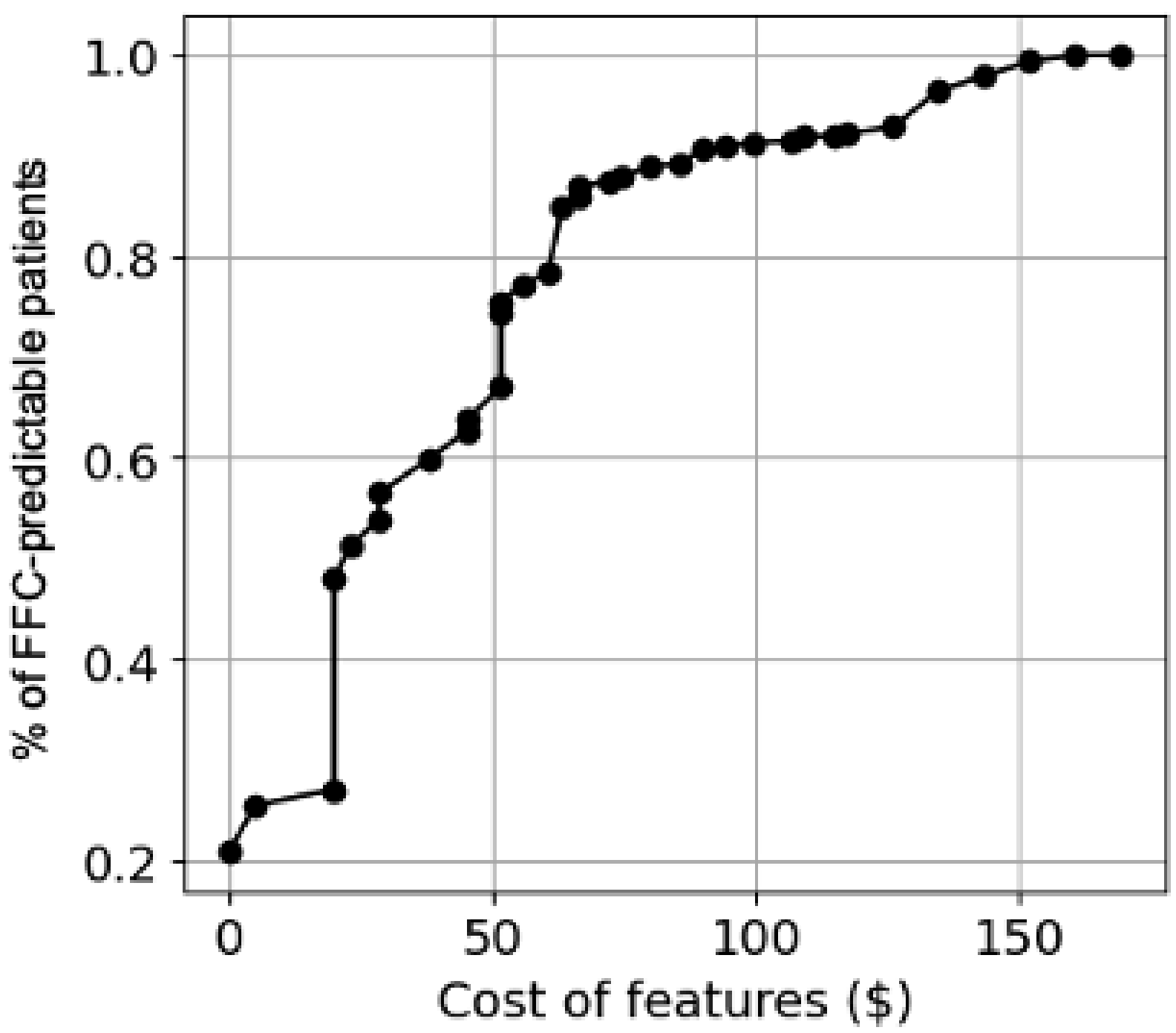

**Figure S3. Cost analysis for the 10-year mortality.** The % of FFC-predictable patients with the increase of monetary cost of features for the NHANES cohort.

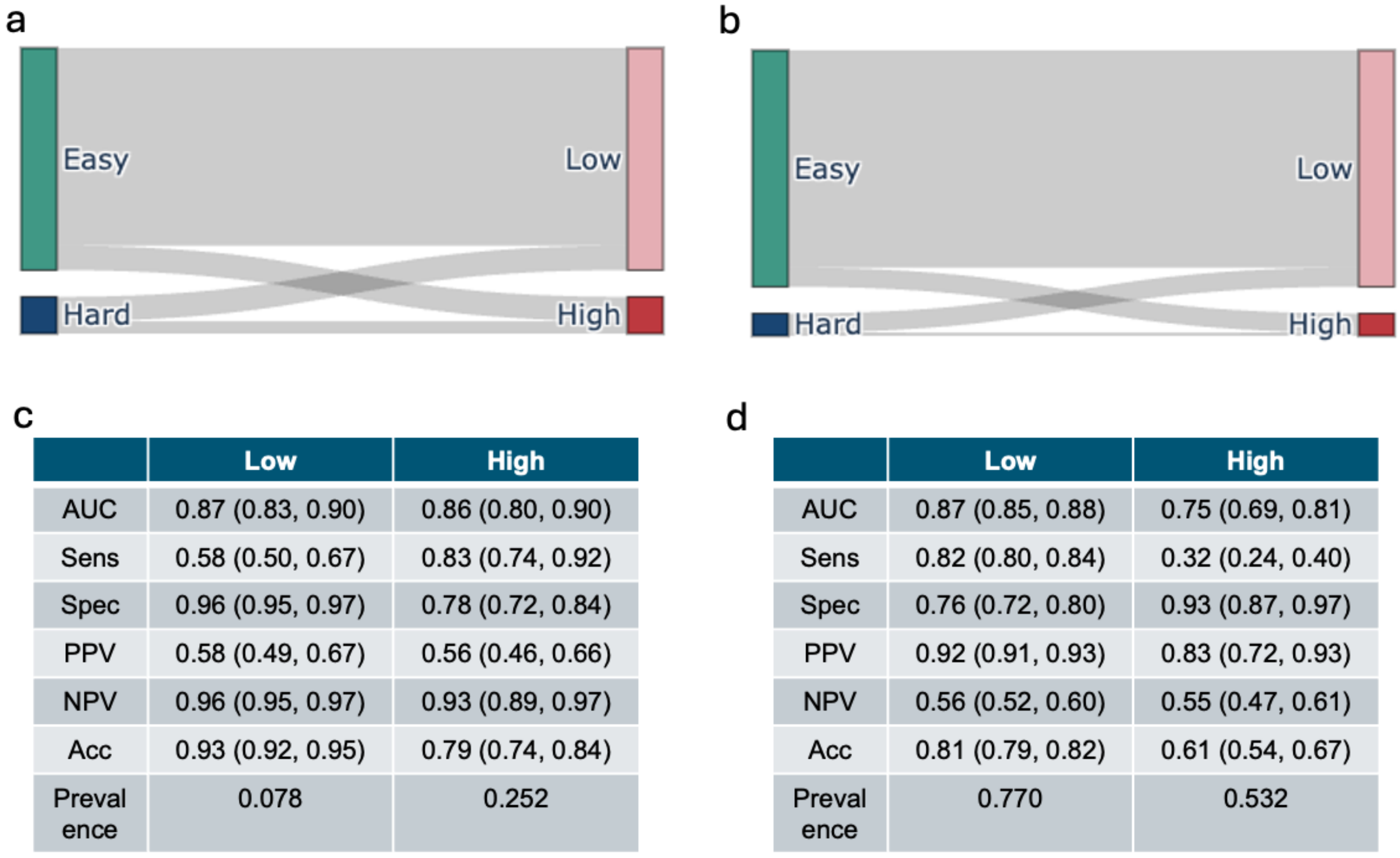

c

| | Low | High |
|---|---|---|
| AUC | 0.87 (0.83, 0.90) | 0.86 (0.80, 0.90) |
| Sens | 0.58 (0.50, 0.67) | 0.83 (0.74, 0.92) |
| Spec | 0.96 (0.95, 0.97) | 0.78 (0.72, 0.84) |
| PPV | 0.58 (0.49, 0.67) | 0.56 (0.46, 0.66) |
| NPV | 0.96 (0.95, 0.97) | 0.93 (0.89, 0.97) |
| Acc | 0.93 (0.92, 0.95) | 0.79 (0.74, 0.84) |
| Prevalence | 0.078 | 0.252 |

d

| | Low | High |
|---|---|---|
| AUC | 0.87 (0.85, 0.88) | 0.75 (0.69, 0.81) |
| Sens | 0.82 (0.80, 0.84) | 0.32 (0.24, 0.40) |
| Spec | 0.76 (0.72, 0.80) | 0.93 (0.87, 0.97) |
| PPV | 0.92 (0.91, 0.93) | 0.83 (0.72, 0.93) |
| NPV | 0.56 (0.52, 0.60) | 0.55 (0.47, 0.61) |
| Acc | 0.81 (0.79, 0.82) | 0.61 (0.54, 0.67) |
| Prevalence | 0.770 | 0.532 |

**Figure S4. The association between FSA-based patient grouping and predictive uncertainty-based patient grouping.** The left figure/table are for prolonged ventilation prediction, right figure/table are for 10-year mortality prediction. (a,b) show how the easy and hard group from FSA-based patient grouping associates with low and high predictive uncertainty groups from forestCI. (c, d) show the performance metrics for low and high predictive uncertainty groups. PPV: positive predictive value, NPV: negative predictive value, Acc: accuracy.

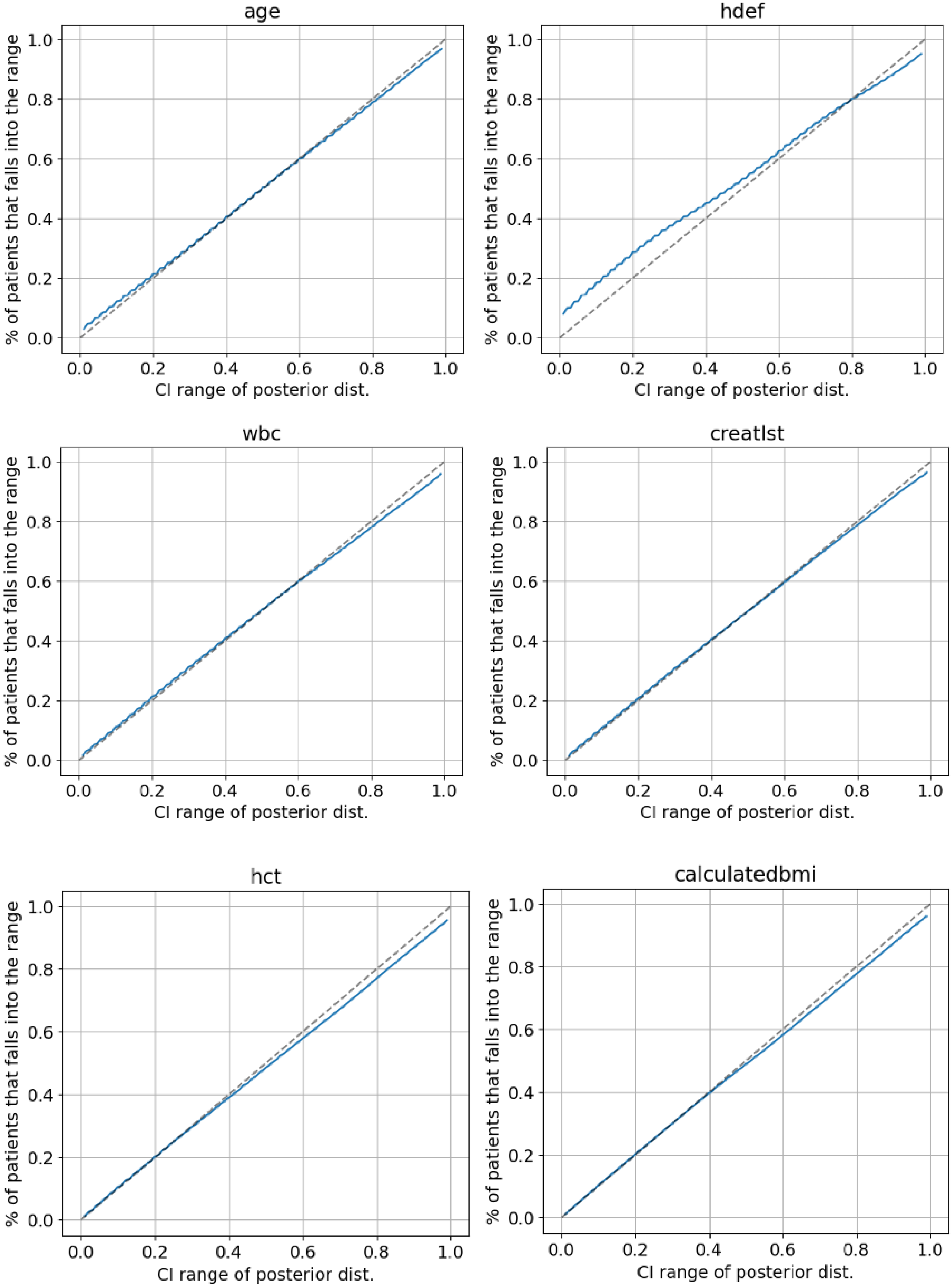
age
hdef
wbc
creatlst
hct
calculatedbmi
% of patients that falls into the range
CI range of posterior dist.
0.0
0.2
0.4
0.6
0.8
1.0

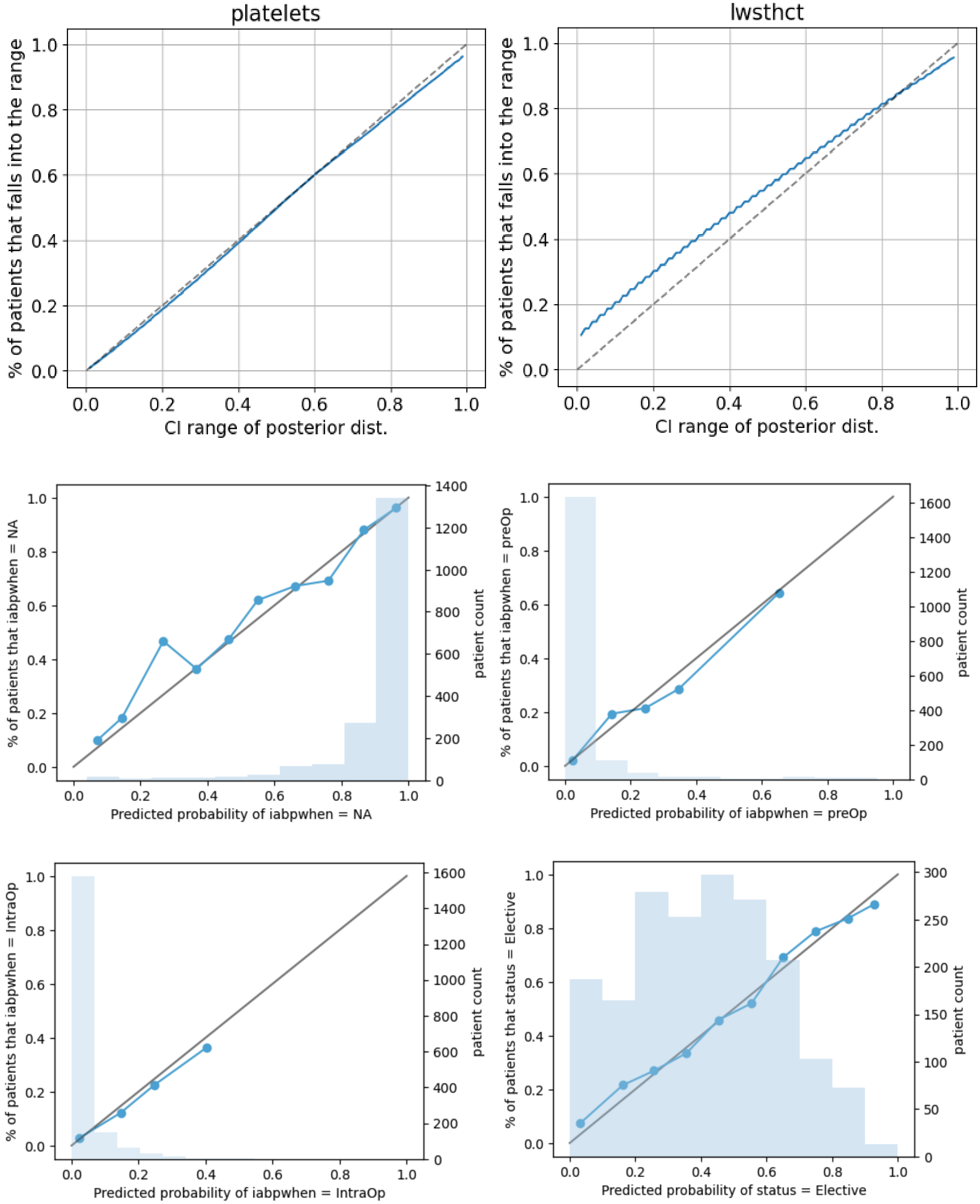

platelets
% of patients that falls into the range
CI range of posterior dist.
lwsthct
% of patients that falls into the range
CI range of posterior dist.
% of patients that iabpwhen = NA
patient count
Predicted probability of iabpwhen = NA
% of patients that iabpwhen = preOp
patient count
Predicted probability of iabpwhen = preOp
% of patients that iabpwhen = IntraOp
patient count
Predicted probability of iabpwhen = IntraOp
% of patients that status = Elective
patient count
Predicted probability of status = Elective

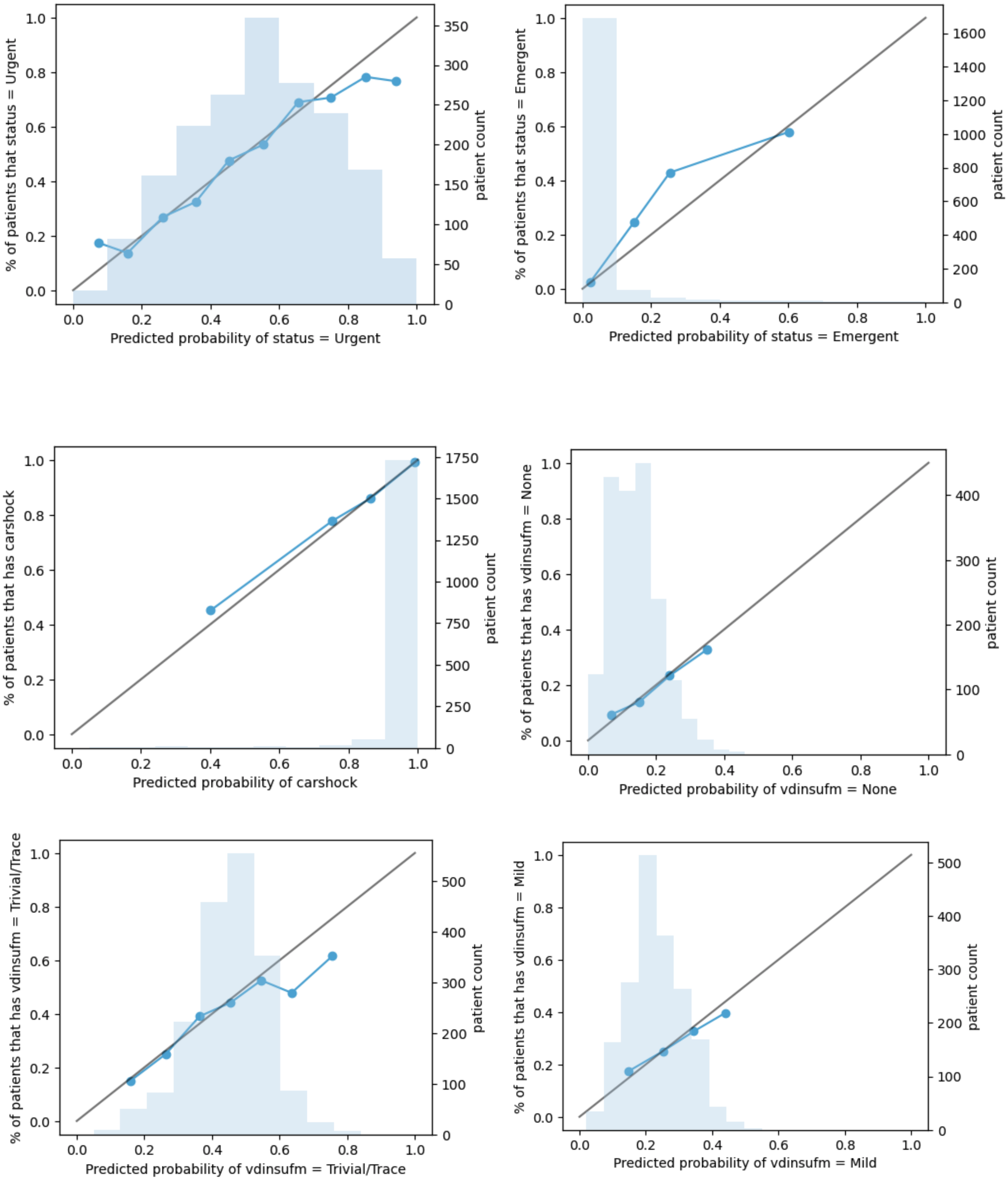
% of patients that status = Urgent
Predicted probability of status = Urgent
patient count
% of patients that status = Emergent
Predicted probability of status = Emergent
% of patients that has carshock
Predicted probability of carshock
% of patients that has vdinsufm = None
Predicted probability of vdinsufm = None
% of patients that has vdinsufm = Trivial/Trace
Predicted probability of vdinsufm = Trivial/Trace
% of patients that has vdinsufm = Mild
Predicted probability of vdinsufm = Mild

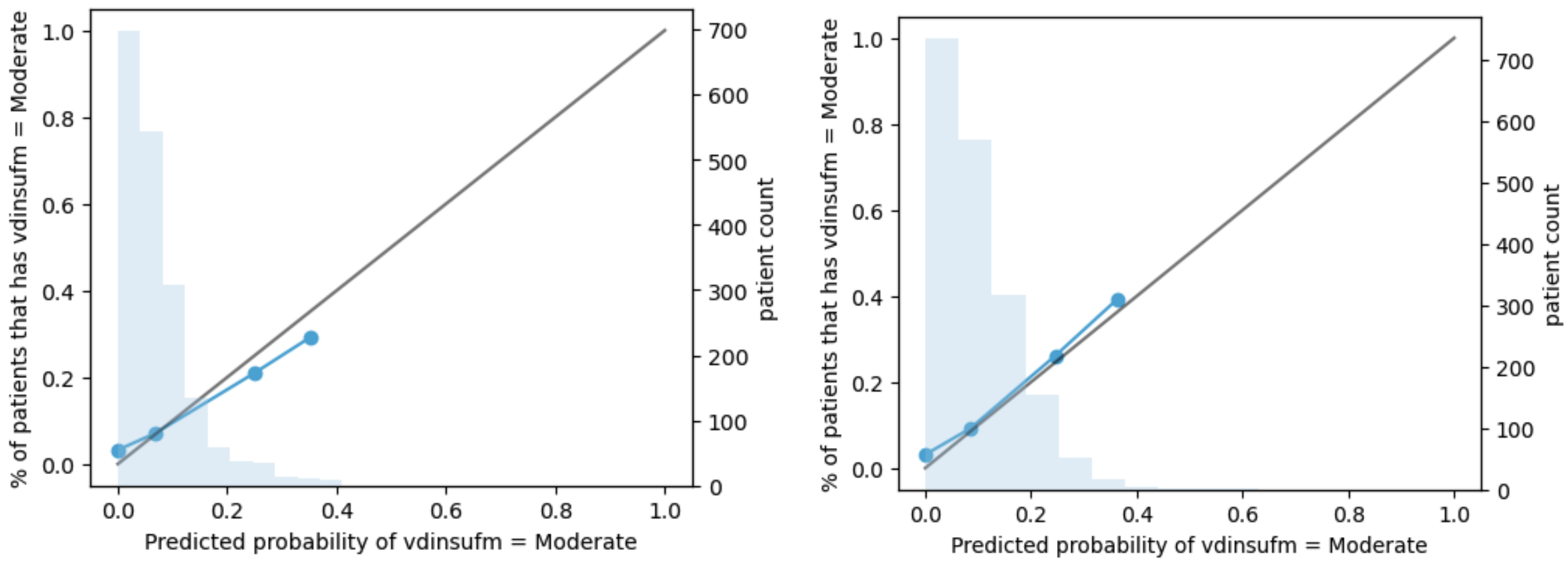

**Figure S5. Validation of the estimated posterior distribution on STS heart surgery patient cohort.** The posterior distribution generated from the ablation study for the feature ranking are examined. For continuous variables (age, hdef, wbc, creatlst, hct, bmi, platelets and lwsthct), the posterior distribution is validated by comparing the credible intervals (x axis) and the percentage of patients falling into the credible intervals. For categorical variables (iabpwhen, status, carshock and vdinsufm), calibration is plotted for each unique value for categorical variables, where x axis is the predicted probability of the value, y axis is the % of patients having this value.

Table S1. 12 features used for developing the baseline ML model for the STS prolonged ventilation prediction.

| Acronyms | Description of variables [values for categorical variables] |
|---|---|
| iabpwhen | When the IABP was inserted<br>[No, Yes, preoperative, Yes intraoperative] |
| creatlst | Indicate the creatinine level closest to the date and time prior surgery |
| age | For age >89, convert it into all into 90. |
| hdef | Ejection fraction |
| bmi | Body mass index |
| status | Clinical status of the patient prior to entering the operating room.<br>[Elective, Urgent, Emergent, Emergent Salvage] |
| hct | Indicate the pre-operative Hematocrit level at the date and time closest to surgery |
| lwsthct | The lowest measured hematocrit recorded in the operating room. |
| carshock | If the patient developed cardiogenic shock. |
| wbc | the pre-operative White Blood Cell (WBC) count closest to the date and time prior to<br>surgery but prior to anesthetic management |
| platelets | Platelet count closest to the date and time prior to surgery but prior to anesthetic management |
| vdinsufm | Whether there is evidence of Mitral valve insufficiency/regurgitation.<br>[None, Trivial/Trace, Mild. Moderate, Severe, Not documented] |

Table S2. 35 features used for developing the baseline ML model for the NHANES mortality prediction. (Features are reproduced from Erion et al.)

| Bun | sgot | Age | Alkaline phosphatase | Band neutrophils |
|---|---|---|---|---|
| Basophils | Calcium | Cholesterol | Creatinine | Eosinophils |
| Height | Hematocrit | Hemoglobin | Lymphocytes | Monocytes |
| Physical activity | Platelets | Potassium | Pulse pressure | Red blood cells |
| Sedimentation rate | Segmented neutrophils | Serum albumin | Serum protein | Sex |
| Sodium | Systolic blood pressure | Total bilirubin | Uric acid | Urine albumin |
| Urine glucose | Urine hematest | Urine pH | Weight | White blood cells |

Table S3. Summary of 5 different feature ranking methods on the STS prolonged ventilation prediction

| features | LR coef. | features | MDI | features | Permutation | Features | SHAP value |
|---|---|---|---|---|---|---|---|
| iabpwhen | 0.831816 | iabpwhen | 0.193228 | iabpwhen | 0.054221 | iabpwhen | 0.062041 |
| lwsthct | 0.647396 | BMI | 0.093717 | lwsthct | 0.004058 | lwsthct | 0.018403 |
| vdinsufm | 0.433178 | wbc | 0.091057 | carshock | 0.003824 | hdef | 0.018197 |
| status | 0.349949 | hdef | 0.090954 | hdef | 0.003373 | creatlst | 0.018077 |
| hdef | 0.290805 | hct | 0.089641 | status | 0.002832 | status | 0.014824 |
| wbc | 0.21306 | creatlst | 0.087943 | vdinsufm | 0.002399 | carshock | 0.014333 |
| carshock | 0.185805 | platelets | 0.085919 | hct | 0.002146 | vdinsufm | 0.013578 |
| BMI | 0.17923 | lwsthct | 0.072722 | wbc | 0.001984 | hct | 0.011563 |
| platelets | 0.159598 | age | 0.072514 | age | 0.001804 | wbc | 0.01106 |
| age | 0.134899 | carshock | 0.044288 | creatlst | -1.80E-05 | BMI | 0.009368 |
| creatlst | 0.116948 | status | 0.040361 | BMI | -0.00018 | age | 0.008481 |
| hct | 0.00945 | vdinsufm | 0.037654 | platelets | -0.0011 | platelets | 0.007191 |

Table S4. Time and monetary cost for 12 features for STS prolonged ventilation prediction

| Testing/Action | Variables | Time cost | Monetary cost |
|---|---|---|---|
| Comprehensive metabolic panel (CMP) | creatlst | 15 mins - 1 hour | $ 277 |
| Complete blood count (CBC) | lwsthct, wbc, hct, platelets | 15 mins - 1 hour | $ 119 |
| Echo ultrasound | vdinsufm, hdef | 1 hour - 4 weeks | $ 573 |
| Demographics | bmi, age | 15s | $ 0 |
| History & Physical | carshock | 15s | $ 0 |
| Procedure | iabpwhen | 15s | $ 0 |
| Chart review | status | 15s | $ 0 |

Table S5. Monetary cost for 35 features for NHANES mortality prediction. (Cost estimates were obtained from the publicly available GitHub repository associated with Erion et al.)

| Feature | Cost | Feature | cost |
|---|---|---|---|
| Bun | 4.39 | Potassium | 5.11 |
| Basophils | 8.63 | Serum albumin | 5.5 |
| Height | 0 | Total bilirubin | 5.57 |
| Physical activity | 0 | Urine pH | 2.41 |
| Sedimentation rate | 3 | Alkaline phosphatase | 9.56 |
| Sodium | 5.35 | Creatinine | 5.69 |
| Urine glucose | 4.37 | Lymphocytes | 8.63 |
| sgot | ? | Pulse pressure | 0 |
| Calcium | 5.73 | Serum protein | 4.07 |
| Hematocrit | 2.63 | Uric acid | 5.02 |
| Platelets | 7.18 | Weight | 0 |
| Segmented neutrophils | 8.63 | Band neutrophils | 8.63 |
| Systolic blood pressure | 0 | Eosinophils | 8.63 |
| Urine hematest | 2.41 | Monocytes | 8.63 |
| Age | 0 | Red blood cells | 3.35 |
| Cholesterol | 14.88 | Sex | 0 |
| Hemoglobin | 2.63 | Urine albumin | 6.42 |
| | | White blood cells | 7.18 |